\documentclass[letterpaper, 10 pt, conference]{ieeeconf}  

\IEEEoverridecommandlockouts                              

\usepackage{amsmath}
\usepackage{amssymb}
\usepackage{amsfonts}
\usepackage{algorithmicx, algpseudocode}
\usepackage{mathtools}
\usepackage{graphicx}
\usepackage{float}
\usepackage{array}
\usepackage{tikz}
\usepackage{listings}
\usepackage{hyperref}
\usepackage{graphicx}
\usepackage{cite}
\usepackage{dsfont}
\usepackage{mathtools,etoolbox}
\usepackage[ruled]{algorithm2e}
\usepackage[none]{hyphenat}
\usepackage[utf8]{inputenc}
\usepackage[english]{babel}
\usepackage{multirow}

\hypersetup{colorlinks=true,urlcolor=blue,citecolor=red}

\newcommand{\etal}{\textit{et al}.}

\DeclareMathAlphabet{\mathpzc}{OT1}{pzc}{m}{it}

\title{\LARGE \bf
SpikeMS: Deep Spiking Neural Network for Motion Segmentation}

\author{Chethan M. Parameshwara, Simin Li, Cornelia Ferm\"uller, Nitin J. Sanket, Matthew S. Evanusa, Yiannis Aloimonos

\thanks{\textit{Chethan M. Parameshwara and Simin Li contributed equally to this work. (Corresponding author: Chethan M. Parameshwara)}} 
\thanks{ All authors are associated with the Perception and Robotics Group, University of Maryland, College Park. Emails: \{{\tt\footnotesize cmparam9, sli12348, fer, nitin, evanusa, yiannis}\} {\tt \footnotesize @umiacs.umd.edu}}} 

\begin{document}

\makeatletter
\g@addto@macro\@maketitle{
\begin{figure}[H]
  \setlength{\linewidth}{\textwidth}
  \setlength{\hsize}{\textwidth}
    \centering
    \includegraphics[width=\textwidth]{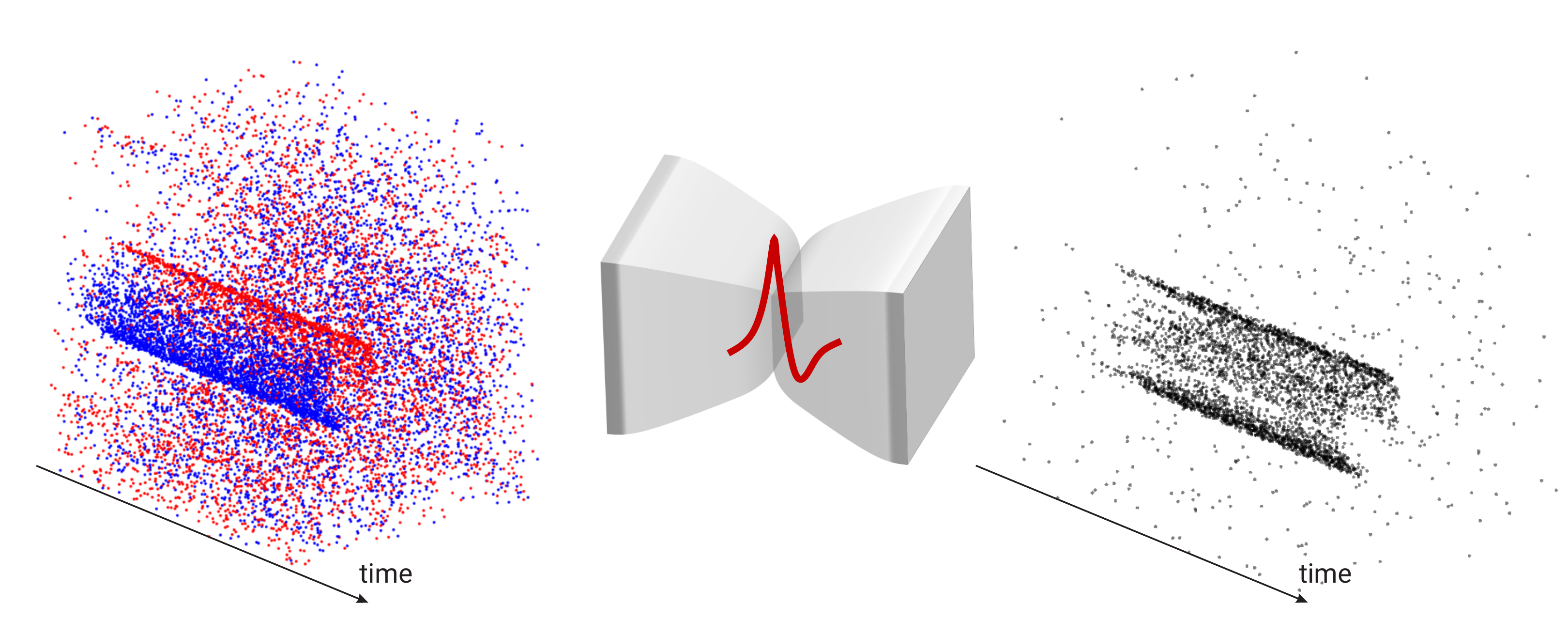}
    \caption{Event-based motion segmentation pipeline using a deep spiking neural network. Left to right: Event stream input represented as red (brightness increase) and blue (brightness decrease), representation of the proposed encoder-decoder spiking neural network called \textit{SpikeMS} and the output predicted spike containing only the region of moving object(s). \textit{All the images in this paper are best viewed in color on a computer screen at a zoom of 200\%.}}
    \vspace{-20pt}
    \label{fig:SpikeMSBanner}
    \end{figure}
}
\maketitle
\thispagestyle{plain}
\pagestyle{plain}

\setcounter{figure}{1}

\begin{abstract}
Spiking Neural Networks (SNN) are the so-called third generation of neural networks which attempt to more closely match the functioning of the biological brain.  They inherently encode temporal data, allowing for training with less energy usage and can be extremely energy efficient when coded on neuromorphic hardware. In addition, they are well suited for tasks involving event-based sensors, which match the event-based nature of the SNN.  However, SNNs have not been as effectively applied to real-world, large-scale tasks as standard Artificial Neural Networks (ANNs) due to the algorithmic and training complexity. To exacerbate the situation further, the input representation is unconventional and requires careful analysis and deep understanding. In this paper, we propose \textit{SpikeMS}, the first deep encoder-decoder SNN architecture for the real-world large-scale problem of motion segmentation using the event-based DVS camera as input. To accomplish this, we introduce a novel spatio-temporal loss formulation that includes both spike counts and classification labels in conjunction with the use of new techniques for SNN backpropagation. In addition, we show that \textit{SpikeMS} is capable of \textit{incremental predictions}, or predictions from smaller amounts of test data than it is trained on. This is invaluable for providing outputs even with partial input data for low-latency applications and those requiring fast predictions. We evaluated \textit{SpikeMS} on challenging synthetic and real-world sequences from EV-IMO, EED and MOD datasets and achieving results on a par with a comparable ANN method, but using potentially 50 times less power.

\end{abstract}


\section*{Supplementary Material}
The accompanying video and supplementary material are available at
\url{prg.cs.umd.edu/SpikeMS}. 

\section{Introduction}

The animal brain is remarkable at perceiving motion in complex scenarios with high speed and extreme energy efficiency. Inspired by the animal brain, an alternative version of Artificial Neural Networks (ANNs) called Spiking Neural Networks (SNNs) aim to replicate the \textit{dynamical system} aspects of living neurons. In contrast to standard ANNs which are essentially networks of complex functions, SNNs are comprised of networks of neurons modeled as differential equations, and inherently encode temporal data and offer low power and highly parallelizable computations. Furthermore, they possess the capability to deliver predictions whose confidence scale with the availability of input data \cite{diehl2015fast,pfeiffer2018deep}. These low-power and low-latency properties are of great use to real-world robotics applications such as self-driving cars or drones, which demand fast responses during navigation in challenging scenarios \cite{sanket2019evdodgenet}. 

Until recently, SNNs have been restricted to simple tasks and small datasets due to instability in learning regimes \cite{gehrig2020event}. 
 Recent development in new spike learning mechanisms \cite{zenke2018superspike, shrestha2018slayer} 
 has made it possible to design SNNs for real-world robotics applications. This coupled with neuromorphic processors such as Intel's $^{\text{\textregistered}}$ Loihi \cite{davies2018loihi} and IBM's TrueNorth\cite{akopyan2015truenorth}) along with neuromorphic sensors such as DVS \cite{lichtsteiner2008128} and ATIS\cite{posch2010qvga}) have made it possible for producing real-world prototypes, drastically enhancing the appeal of such technologies.
  
 In this work, we propose a deep SNN architecture called \textit{SpikeMS} for the problem of motion segmentation using a monocular event camera. We consider the data from event sensors as they are well-suited for motion segmentation (due to the disparity in event density at object boundaries) and are a natural fit for SNNs (due to their temporal nature). We will now formally define the problem statement and our main contributions.

\subsection{Problem Formulation and Contributions}



We address the following question: \textit{How do you learn to segment
the scene into background and foreground (moving objects) using a Spiking Neural Network from the data of a moving monocular event camera?}


Our spiking neural network, \textit{SpikeMS}, takes the event stream as input and outputs predictions of each event's class association as either foreground (moving object) or background (moving due to camera motion).


The model learns to distinguish between the spatio-temporal patterns of moving objects and the background. To the best of our knowledge, this is the first end-to-end deep encoder-decoder SNN. In particular, we evaluate our network on the task of motion segmentation using event input. 

The main contributions of the paper are given below: 
\begin{itemize}
    \item A novel end-to-end deep encoder-decoder Spiking Neural Network (SNN) framework for motion segmentation from event-based cameras.
    \item Demonstration of ``early'' evaluation of the network (at low latency), which  we  call \textit{Incremental Predictions}, for imprecise but fast detection of moving objects for variable-sized integration windows. 
\end{itemize}



\section{Related Work}

\subsection{Spiking Neural Network Weight Learning Rules}
While the concept of a spiking neuron has been around for a few decades \cite{hodgkin1952quantitative}, their progress has been bounded by the difficulty in training due to the ubiquitous vanishing gradient problem for deep neural networks.
In SNNs, the neurons output pulses that are non-differentiable, rendering attempts at directly applying the backpropagation algorithm non-trivial.  Early attempts at training SNNs revolved around more biologically plausible Hebbian-style mechanisms \cite{sejnowski1989hebb} that only involve \textit{local} updates, such as \textit{Spike Time Dependent Plasticity} (STDP) \cite{nessler2009stdp}, avoiding gradient issues. Work in this field continues to this day, with results \cite{kheradpisheh2018stdp,evanusa2020deep, lee2018training} demonstrating utilities of STDP in training deep SNNs. 
Early attempts at incorporating backpropagtion into SNNs involved first training a traditional ANN, and then transferring the learned weights to an SNN \cite{diehl2015fast}. Recent methods, which we build off of here, allow the SNN to be directly trained through backpropagation by finding a surrogate, continuous value function that roughly correlates for the spike activity \cite{lee2016training, zenke2018superspike, shrestha2018slayer}.

\subsection{SNNs for Visual Tasks and Event Data}
There has been a renewed interest in using SNNs to process data directly from event-based visual sensors, such as the DVS since the sensor produces spike-like activity that fits well with SNN neurons. Applications of SNNs in this domain include classification problems \cite{neil2016learning} such as digit recognition \cite{orchard2015converting}, object recognition \cite{orchard2015hfirst} gesture recognition\cite{amir2017low}, and optical flow\cite{haessig2018, paredes2019unsupervised}. Recent development of neuromorphic processors such as the Intel Loihi \cite{davies2018loihi} has lead to the deployment of SNNs on hardware \cite{haessig2018, haessig2019spiking, renner2019event}. 

Closest related to our work, in \cite{gehrig2020event} recently a neural architecture of multiple layers has been designed. A six-layer neural network (five convolutional and one pooling layer) is used to learn with supervision to regress the three parameters of camera rigid rotation. In Lee \etal \cite{lee2020spike} a deep hybrid encoder decoder architecture was designed for self-supervised optic flow estimation. The encoding layers are SNN with the  backpropagation learning employing the  approximation of \cite{lee2016training}, and the residual and decoding layers are ANN with the self-supervised loss computed from the images of a  combined DVS and image sensor (DAVIS).  Our network is the first architecture for the problem of motion segmentation with event data.

  Event-based cameras have been recognized as a promising sensor for the problem of segmentation and detection of independently moving objects, as the event stream carries essential information about the movement of object boundaries\cite{barranco2018real}. Classical approaches \cite{mitrokhin2018event, stoffregen2019event, parameshwara2021moms} treat motion segmentation as a geometric problem and model it as an artifact of motion compensation of events. In ANN approaches, the input representation is formed by binning the events within a time-interval and convert to an image-like frame based  structure\cite{mitrokhin2019ev, sanket2019evdodgenet} the so called ``event-frames". Our approach is similar to \cite{mitrokhin2020learning}, but rather than creating event-frames, a sampled version of the event stream is fed directly into the network, taking advantage of the SNN's temporal nature in conjunction with the temporal nature of the event stream.

\section{SpikeMS Architecture}
\begin{figure}[t!]
\begin{center}
    \includegraphics[width=\columnwidth]{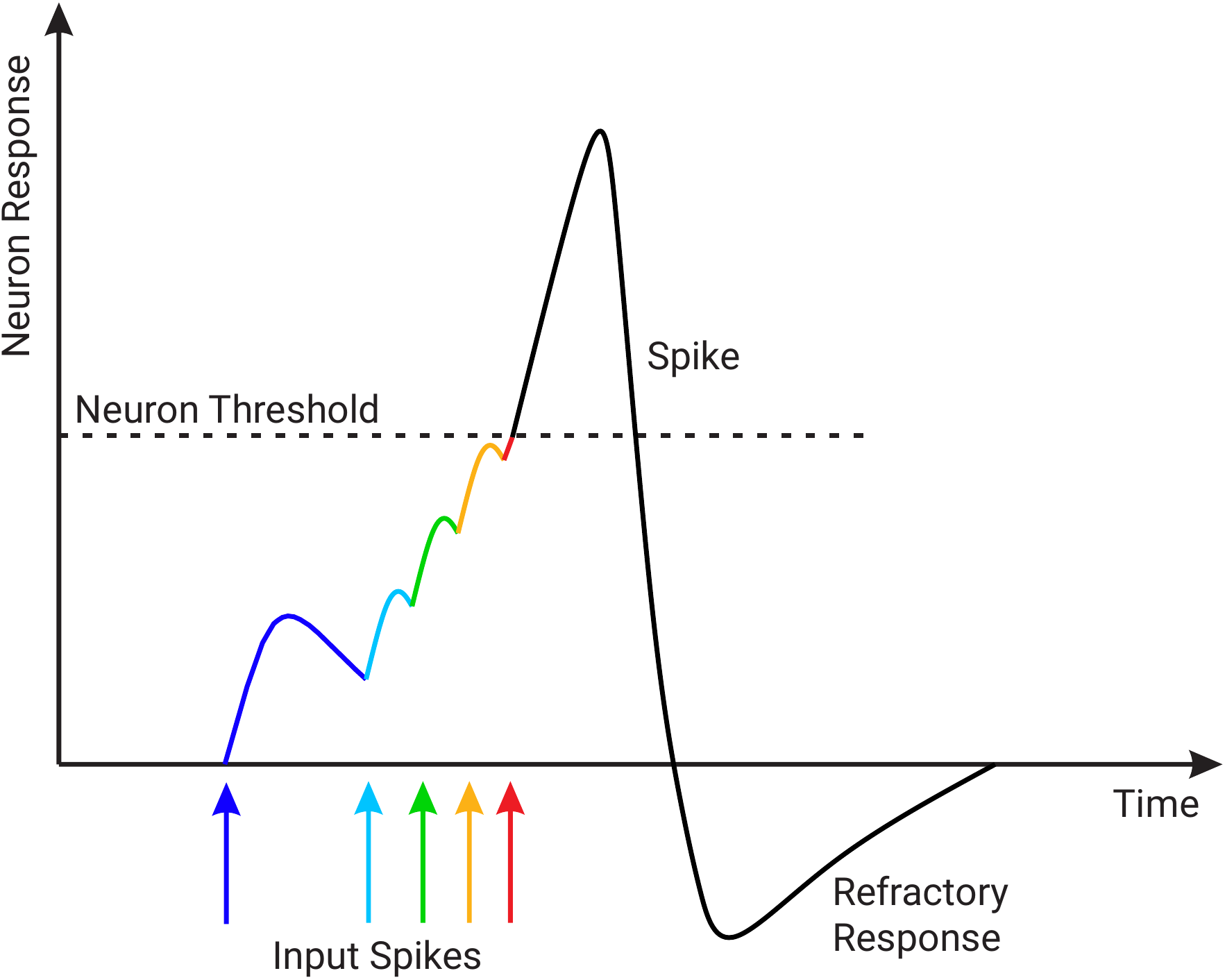}
\end{center}
\caption{Depiction of the dynamical activity of a spiking neuron.  The neuron receives input coming either from the data or lower layers (shown here as colored arrows), which generate bumps in the membrane voltage; we refer to this voltage in the paper as $u(t)$.  If the voltage $u(t)$ exceeds a threshold $\vartheta$, shown here as the dotted line, the neuron outputs a spike, and then enters a refractory phase where it is less likely to fire another spike for a short time.  Computationally, this spiking after passing a threshold amounts to feeding $u(t)$ through the spike function $f_{s}$. The effect that incoming pulses have on the voltage, and the extent of the refractory response, is governed in the Spike Response Model (SRM) \cite{gerstner1995time} via the $\varepsilon$ and $\nu$ kernels respectively (See Section \ref{sec::SRM_Model} for more detail).} 
\label{fig:spiking_neuron}
\end{figure} 

\subsection{Event Camera Input}
A traditional camera records frames at a fixed frame rate by integrating the number of photons for the chosen shutter time for all pixels \textit{synchronously} (in a global shutter camera). In contrast, an event camera only records the polarity of logarithmic brightness changes \textit{asynchronously} at each pixel, resulting in asynchronous packets of information containing the pixel location and the time of the change known as an \textit{event}. If the brightness at time $t$ of a pixel at location $\mathbf{x}$ is given by $I_{t,\mathbf{x}}$ an event is triggered when 

\begin{equation}
    \Vert \log\left( I_{t+\delta t,\mathbf{x}}\right) - \log\left( I_{t,\mathbf{x}}\right) \Vert_1 \ge \tau
\end{equation}

where $\delta t$ is a small time increment and $\tau$ is a trigger threshold. Each event outputs the following data:  $\mathbf{e} = \left\{\mathbf{x}, t, p \right\}$, where $p = \pm 1$ denotes the sign of the brightness change. We will denote the event stream in a spatio-temporal window as $\mathcal{E}\left( t, t+\delta t\right) = \{e_i\}^N_{i=1}$ ($N$ is the number of events). We refer to it as event slice/stream/cloud/volume or spike train interchangeably.

\subsection{Spiking Neuron Model}
\label{sec::SRM_Model}

Spiking Neurons, unlike traditional \textit{rate-encoding} neurons (commonly used neurons in standard ANNs), implicitly encode time in their formulation.  They are modeled loosely after neurons in the brain, following the pioneering work by Hodgkin-Huxley \cite{hodgkin1952quantitative} which laid the groundwork for differential equation modeling of neuronal activity.  
We utilize the Spike Response Model (SRM) which similar to all spiking neuron models, sums up incoming voltage from pre-synaptic neurons, but contains two filters: a filter that accounts for the neuron's \textit{self-refractory} response denoted as $\nu$, and a \textit{spike response kernel} that accounts for the integration of incoming pre-synaptic pulses denoted as $\varepsilon$. For a given neuron $i$ at  timestep $t$, the update of the neuron's synaptic potential dynamics takes the form of:

\begin{equation}
\begin{aligned}
    u_{i}(t) & = \Big(\sum_j w_{j} (\varepsilon * s_{j})\Big) + (\nu * s) \\
    & = \textbf{w}^{\top} \textbf{a} + (\nu * s)
\end{aligned}
\end{equation}

for all incoming weight connections from pre-synaptic neurons $1,...,j$, where $a(t) = (\varepsilon * s)(t)$, $s_i(t)$ is an input spike train in a neuron and $*$ denotes the convolution operator. An output spike is generated whenever $u(t)$ reaches the spiking threshold $\vartheta$ (the dotted line in Fig. \ref{fig:spiking_neuron})



The motivation for using SRM neuron types is that it inexpensively models the refractory behavior of neurons without having to run multiple differential equation solvers, as seen in other models. 


Specific choices of $\varepsilon$ and $\nu$ reduce the SRM equations to a LIF neuron \cite{Gerstner:2008}.  Here, we use the formulation from \cite{gehrig2020event}:

\begin{equation}
    \varepsilon(t) = \frac{t}{\tau_s} e^{1-\frac{t}{\tau_s}}\mathcal{H}(t)   
\end{equation}

\begin{equation}
    \nu(t) = -2\vartheta e^{1-\frac{t}{\tau_r}}\mathcal{H}(t)   
\end{equation}

where $\mathcal{H}$ is the Heaviside function, and $\tau_s$ and $\tau_r$ are the spike response and refractory time constants.  

The activity of the neurons is then propagated forward through the layers of the network, in the same manner as an ANN.  The feed-forward weight matrix $\textbf{W}^{(l)}$ $= [\textbf{w}_{1}, ... , \textbf{w}_{N_{l+1}}]$ for a given layer $l$ with $N_{l}$ neurons is applied to the activity resulting from the spike response kernel, added to the refractory activity and then thresholded. Thus, for all layers $l$ in the network, the activity is forward-propagated as:
\begin{equation}
    a^{(l)}(t) = (\varepsilon_{d} * s^{(l)})(t)
\end{equation}
\begin{equation}
        u^{(l+1)}(t) = \textbf{W}^{(l)} a^{(l)}(t) + (\nu * s^{(l+1)}(t))
\end{equation}
\begin{equation}
    s^{(l+1)}(t) = f_{s}(u^{(l+1)}(t))
\end{equation}

where $f_s$ is the thresholding function, $\textbf{W}^{(l)}$ is the forward weight matrix for layer $l$, and $\varepsilon_{d}$ is the spike response kernel with delay, as in \cite{shrestha2018slayer}.  The input to the network, $s^{(0)}$, is the event data over the learning window.

\subsection{Network Architecture}
\label{subsec:architecture}
\textit{SpikeMS} utilizes an end-to-end deep Spiking Neural Network, in contrast to many recent models \cite{lee2020spike} that use a hybrid combination of spiking and rate-encoding layers. To the best of our knowledge, \textit{SpikeMS} is the first end-to-end spike trained deep encoder-decoder network for large scale tasks such as motion segmentation. 



\textit{SpikeMS} consists of a traditional hourglass-shaped layer structure of an autoencoder, with larger layers progressively encoded to smaller representations, which are then decoded back to the original size. 
We use three encoder layers followed by three decoder layers. The first three convolutional layers perform spatial downsampling with a stride of 2 and kernel size of 3x3. The output of the first encoder layer contains 16 channels, and each encoder layer after doubles the number of channels. The last three decoder layers perform spatial upsampling using transposed convolution, with a stride of 2 and kernel size of 3x3. Decoder layers 4 and 5 each halve the number of channels. The last layer (6) outputs the predicted spikes of the moving object(s) using 2 channels, representing positive and negative event polarities.   


\subsection{Spatio-Temporal Loss}
\label{sec::loss}
We propose a novel loss formulation which takes full advantage of the spatio-temporal nature of the event data. Our loss function consists of two parts: a binary cross entropy loss $\mathcal{L}_{\text{bce}}$ and spike loss $\mathcal{L}_{\text{spike}}$. 

The binary cross-entropy loss $\mathcal{L}_{\text{bce}}$ is computed by comparing the predicted temporal spike projection to the ground truth temporal spike projection.  
\begin{equation}
      \mathcal{L}_{\text{bce}} = - \left( \mathds{1}_f\log\left(\hat{\mathpzc{E}}_f\right) + \mathds{1}_b\log\left(\hat{\mathpzc{E}}_b\right)\right)
\end{equation}

Where, the spike projection $\mathpzc{E}$ is obtained as: $\mathpzc{E}\left( \mathbf{x}\right) = \sum_{t}\mathcal{E}\left( \mathbf{x}\right)$. Such a projection converts a spike train into a real-valued output, encoding the frequency of spikes. And the groundtruth foreground and background labels are denoted as $\mathds{1}_f$ and $\mathds{1}_b$ respectively. 



The spike loss $\mathcal{L}_{\text{spike}}$ is derived from the Van-Rossom distance\cite{vanRossum} and measures the distance between two binary spike trains \cite{Kreuz:2011}. $\mathcal{L}_{\text{spike}}$ preserves the temporal precision of the event stream. The ground truth spike labels are generated by applying a binary mask to the event cloud input $\mathcal{E}$, i.e., masking all non-moving-object events (background events) as 0, and keeping intact the events that correspond to the moving object. $\mathcal{L}_{\text{spike}}$ is given by 
\begin{equation}
     \mathcal{L}_{\text{spike}} = \sum_{t = 0}^{t + \delta t} \left( \mathcal{\hat{E}}\left(t, t+\delta t\right)\circ\mathds{1}_f - \mathcal{\tilde{E}}\left(t, t+\delta t\right)\right)^2 \,dt 
\end{equation}

where $\circ$ denotes the Hadamard product and $\mathds{1}_f$ denotes the mask of foreground spikes. 

The overall loss $\mathcal{L}_{\text{total}}$ is given by

\begin{equation}
     \mathcal{L}_{\text{total}} = \mathcal{L}_{\text{bce}} + \lambda\mathcal{L}_{\text{spike}} 
\end{equation}
where $\lambda$ is a weighting factor. The error is backpropagated through the network using SLAYER \cite{shrestha2018slayer}.

\begin{figure}[t!]
\begin{center}
    \includegraphics[width=\columnwidth]{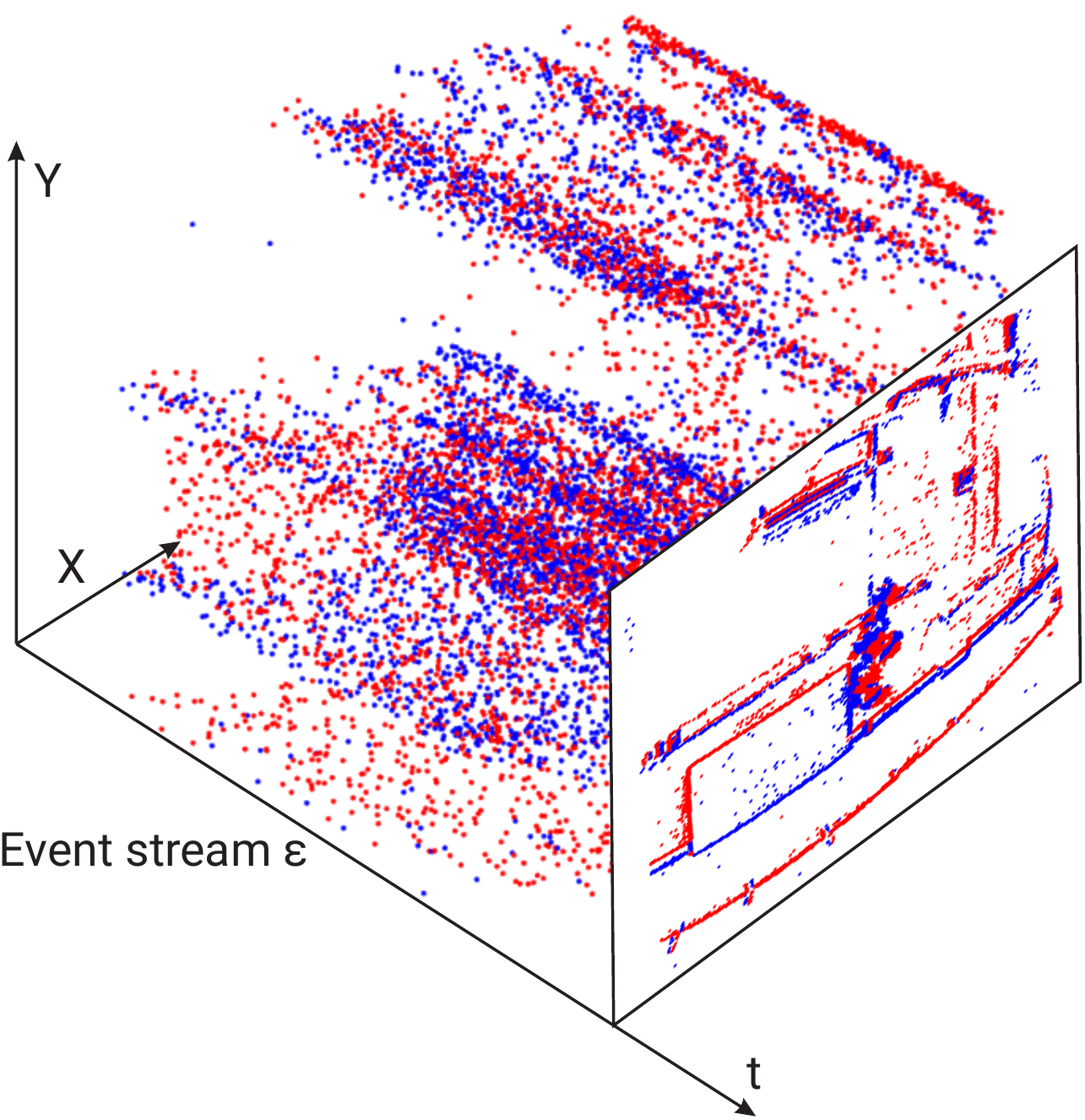}
\end{center}
  \caption{Representation of event stream $\mathcal{E}$ and its corresponding projection (event frame). Note that, only event streams are fed to \textit{SpikeMS}. The event frame is shown only for clarity purposes.} 
\label{fig:dataset}
\end{figure} 

\subsection{Simulation of SNNs on GPU}
SNNs are continuous dynamical systems which can process input event streams asynchronously. However, for our experiments, we simulate the SNN network on a GPU. To achieve this, we need to discretize the event data at fixed time steps. 
To balance the trade-off between accuracy and resource availability \cite{gehrig2020event}, we restrict the simulation time step to one millisecond. We train our SNNs with fixed simulation time window/width/steps $\Delta t_{\text{train}}$ of 10ms for all experiments. However, to test the out-of-domain temporal performance, we test our predictions on simulation time steps $\Delta t_{\text{test}}$ of 1ms to 25ms. To fit the event inputs into fixed time steps, the multiple events with the same polarity and spatial location within a timeframe are represented as a single binary event. This downsampling collapses all events within the window into a single event. The simulated network is trained with the publicly available PyTorch implementation of SLAYER \cite{shrestha2018slayer}.

\section{Experiments and Results}
\begin{figure*}[t!]
\begin{center}
    \includegraphics[width=\textwidth]{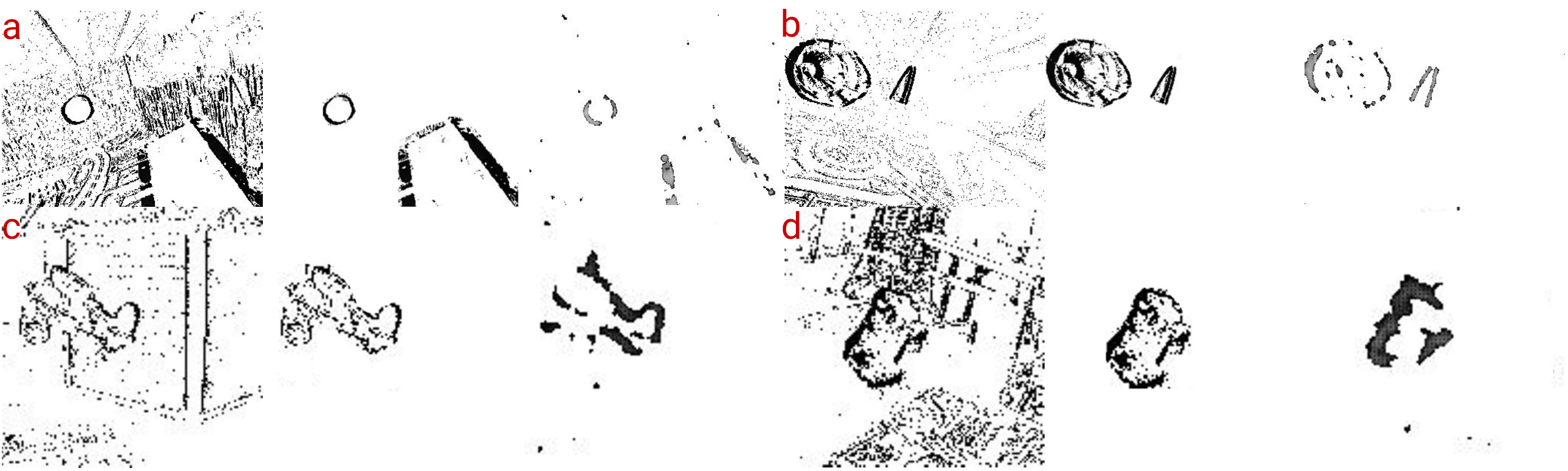}
\end{center}
\caption{Qualitative Evaluation of our approach on two datasets. Top row (a and b): MOD dataset, Bottom row (c and d): EV-IMO dataset. Each sample includes, (left to right) event stream, groundtruth, and network prediction. Here, we show event projections for clarity purposes but \textit{SpikeMS} predicts spatio-temporal spikes.} 
\label{fig:DatasetEval}
\end{figure*} 

We evaluate our approach on publicly available synthetic and real datasets. We demonstrate performance of \textit{SpikeMS} both qualitatively and quantitatively by employing the Intersection over Union ($IoU$) and Detection Rate (DR) metrics \cite{parameshwara2021moms}.

\subsection{Overview of Datasets}
We use the publicly available MOD\cite{sanket2019evdodgenet} and EV-IMO\cite{mitrokhin2019ev} datasets for training and evaluating the motion segmentation predictions. MOD \cite{sanket2019evdodgenet} is a synthetic dataset specifically targeted for learning based motion segmentation approaches. The simulated data contains objects moving in an indoor room-like environment with randomized wall textures, static/dynamic objects and the object/camera trajectories. EV-IMO \cite{mitrokhin2019ev} contains monocular event-based camera data captured in a lab environment with challenging scenarios (multiple objects moving in random trajectories and varying speeds). EV-IMO contains five different sequences (\texttt{boxes}, \texttt{floor}, \texttt{wall}, \texttt{table}, and  \texttt{fast}) which were collected using the DAVIS 346 camera.



\subsection{Quantitative Results}
\begin{table*}[ht]
    \centering
     \caption {Quantitative Evaluation using IoU ($\%$) $\uparrow$ metric on EV-IMO and MOD datasets.}
    \label{tab:compare}
    \resizebox{\textwidth}{!}{
    \begin{tabular}{llllllllllllll}
    \hline
         \multirow{2}{*}{Method} & 
         
        
         \multicolumn{10}{c}{EV-IMO} &  \multicolumn{2}{c}{\multirow{2}{*}{MOD}}\\
         & \multicolumn{2}{c}{\texttt{boxes}}
         & \multicolumn{2}{c}{\texttt{floor}}
         & \multicolumn{2}{c}{\texttt{wall}} 
         & \multicolumn{2}{c}{\texttt{table}} 
         & \multicolumn{2}{c}{\texttt{fast}}  
         & \\
         \hline
         & 100 & 20 & 100 & 20 & 100 & 20 & 100 & 20 & 100 & 20 & 100 & 20 \\
         \hline
         EV-IMO$^\dagger$\cite{mitrokhin2019ev}  & \multicolumn{2}{c}{70$\pm$5}
         & \multicolumn{2}{c}{\textbf{59$\pm$9}}
         & \multicolumn{2}{c}{78$\pm$5} 
         & \multicolumn{2}{c}{79$\pm$6} 
         & \multicolumn{2}{c}{67$\pm$3} 
         & \multicolumn{2}{c}{-} \\
         
         EVDodgeNet\cite{sanket2019evdodgenet}  & \multicolumn{2}{c}{\textbf{67$\pm$8}}
         & \multicolumn{2}{c}{61$\pm$6}
         & \multicolumn{2}{c}{72$\pm$9} 
         & \multicolumn{2}{c}{70$\pm$8} 
         & \multicolumn{2}{c}{60$\pm$10}  
         & \multicolumn{2}{c}{75$\pm$12} \\
         \hline
         GConv$^\dagger$\cite{mitrokhin2020learning} & 81$\pm$8 & 60$\pm$18 & 79$\pm$7 & 55$\pm$19 & 83$\pm$4 & 80$\pm$7 & 57$\pm$14 & \textbf{51$\pm$16} & 74$\pm$17 & \textbf{39$\pm$19} & - & -  \\
         PointNet++\cite{qi2017pointnet} & 71$\pm$22 & 80$\pm$15 & 68$\pm$18 & 76$\pm$10 & 75$\pm$19 & 74$\pm$20 & 62$\pm$28 & 68$\pm$23 & 24$\pm$10 & 20$\pm$6 & 74$\pm$13 & \textbf{67$\pm$15} \\
         
         \hline
        Ours ($\mathcal{L}_{bce}$) & 57$\pm$11 & 59$\pm$7 & 56$\pm$9 & 46$\pm$12 & 62$\pm$8 & 62$\pm$9 & 51$\pm$12 & 45$\pm$12 & 42$\pm$13 & 36$\pm$13 & 62$\pm$11 & 63$\pm$7 \\
        Ours ($\mathcal{L}_{spike}$) & 45$\pm$4 & 52$\pm$7 & 49$\pm$8 & 44$\pm$6 & 53$\pm$15 & 47$\pm$11 & 43$\pm$15 & 37$\pm$4 & 41$\pm$6 & 35$\pm$4 & 55$\pm$11 & 55$\pm$8 \\
        Ours ($\mathcal{L}_{bce} + \mathcal{L}_{spike}$) & 61$\pm$7 & \textbf{65$\pm$8 }& \textbf{60$\pm$5} & 53$\pm$16 & 65$\pm$7 & 63$\pm$6 & 52$\pm$13 & \textbf{50$\pm$8} & 45$\pm$11 & \textbf{38$\pm$10} & 68$\pm$7 & \textbf{65$\pm$5}\\
         \hline
        
    \end{tabular}}
    \tiny{ $^\dagger$Results taken directly from \cite{mitrokhin2020learning}}
\end{table*}



We compare our method against state-of-the-art ANNs (both 2D and 3D) and the results are given in Table \ref{tab:compare}. In particular, 2D ANNs (EV-IMO \cite{mitrokhin2019ev}, EVDodgeNet \cite{sanket2019evdodgenet}) are trained with inputs consisting of event-frames computed by accumulating (or projecting) events on a plane. In contrast, 3D ANNs (GConv \cite{mitrokhin2020learning} and PointNet++ \cite{qi2017pointnet}) are trained directly on the event cloud $\mathcal{E}$.  We evaluate ANN-2D with event frames integrated with a time width $\Delta t$ of 25ms. ANN-3D approaches are evaluated with two time widths $\Delta t$ of 20ms and 100ms similar to \cite{mitrokhin2020learning}. 

During evaluation, \textit{SpikeMS} is tested at $\Delta t_{\text{test}}$ = 100ms and  $\Delta t_{\text{test}}$ = 20ms (trained at $\Delta t_{\text{train}}$=10ms)  for a fair comparison with ANNs-3D. Table \ref{tab:compare} provides the mean $IoU$ results on multiple sequences of the EV-IMO and MOD datasets. We observe that the performance of \textit{SpikeMS} is comparable to ANN-2D and ANN-3D approaches in all cases. However, note that the ANN-2D and ANN-3D perform better in the domain they are trained in as compared to \textit{SpikeMS} and we speculate this is because of more stable training procedures in ANNs. This points to a direction of future work for SNNs of proposing better training methodologies.

In Table \ref{tab:compare}, we also compare our results when trained on different loss functions. We observe that the proposed spatio-temporal loss formulation performs better than just using the spike loss or crossentropy loss as it utilizes the information from both spatial and time domains together.  

Finally, we also compare \textit{SpikeMS} with classical hand-crafted methods in Table  \ref{tab:DetectionRateResults} and we see that, our SNN approach outperforms most hand-crafted methods whilst being deployable directly on neuromorphic hardware. This would lead to huge power savings when deployed on a robot. 





         

\subsection{Incremental Prediction}
\label{sec::incremental}


We test the network's capability to perform \textit{incremental predictions} evaluating the network at different testing discretized window sizes, with $\Delta t_{\text{test}}$ ranging from 1ms to 25 ms, while keeping the training window fixed at $\Delta t_{\text{train}}$ of 10ms. This experiment tests for the out-of-domain generalization performance of \textit{SpikeMS}.

\begin{figure}[t!]
\begin{center}
    \includegraphics[width=\columnwidth]{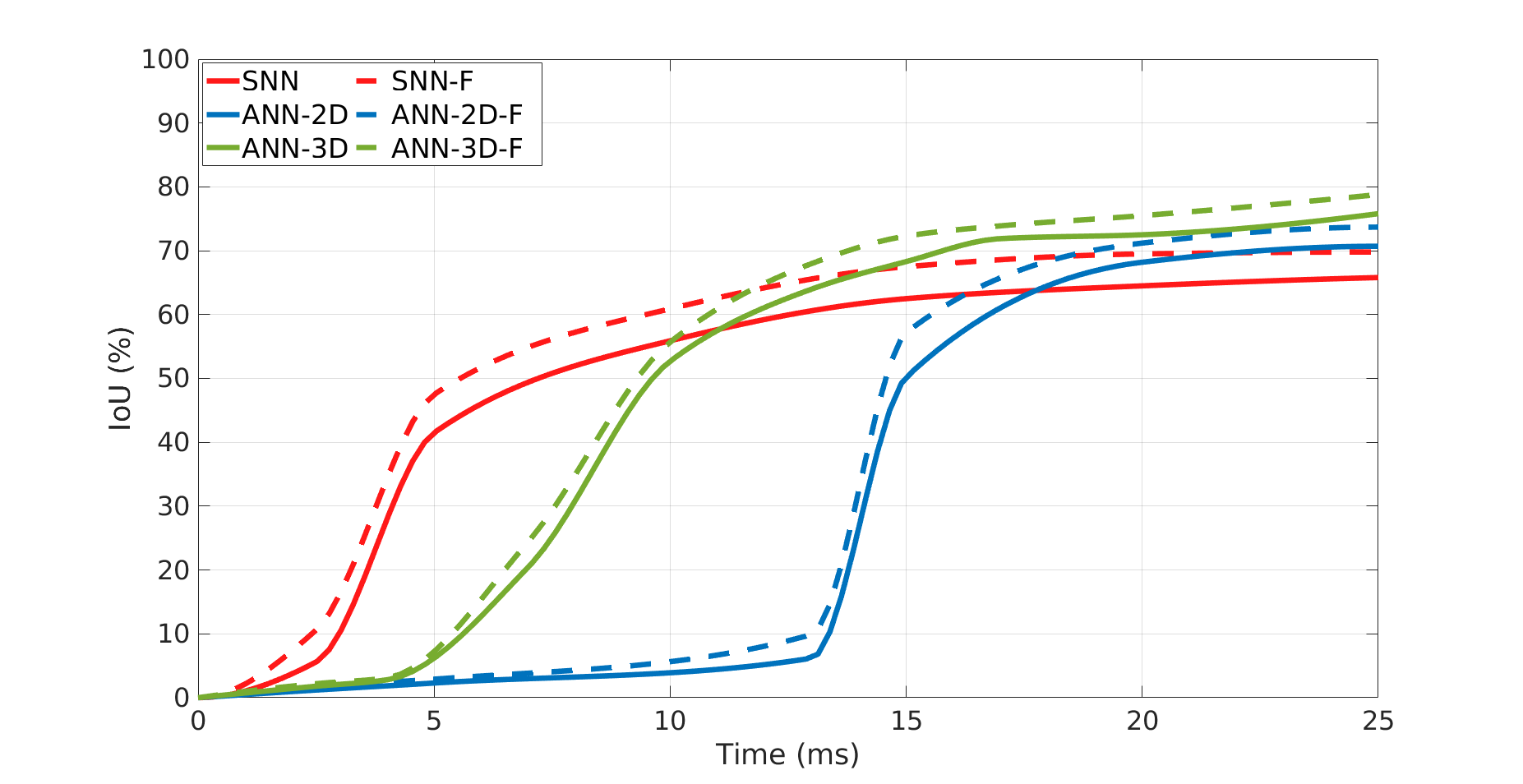}
\end{center}
   \caption{Incremental Prediction: Segmentation accuracy vs. input spike window length in milliseconds for various simulation time width $\Delta t$. SpikeMS is able to achieve good accuracy significantly faster than ANNs, given smaller input data. The dashed lines represent accuracy improvement after employing a filtering (See Sec. \ref{sec::incremental}).} 
\label{fig:incremental}
\end{figure}

All predictions made at $\Delta t < 10$ms can be considered as \textit{incremental predictions} (See Fig. \ref{fig:incremental}), since the testing window is smaller than the training window. This is particularly important for robotics applications since one can filter these incremental predictions to get close to the accuracy of the model with long time predictions but with a lower latency. For example, we can filter predictions (we use a linear Kalman filter \cite{kalman1960new} for filtering) of 3ms to obtain up to $\sim$64\% of the accuracy of 10ms predictions, but with 70\% less latency which might be required for time-critical controllers. We also experiment with values greater than 10ms, where we examine whether longer integration windows yield more accurate results.

Fig. \ref{fig:incremental} shows the plot of accuracy (IoU) versus the duration of input spikes during simulation, considered for prediction. We observe that the prediction accuracy increases over time with the occurrence of more spikes, but critically, that the SNN is able to output reasonable predictions from less spikes. As shown in SNNs outperform ANN-2D and ANN-3D at early stages with less amount of data. We observe that ANN-3D outperforms ANN-2D since it is trained with temporal augmentation techniques as proposed in \cite{mitrokhin2020learning}. Note that the SNN does not rely on temporal augmentations for incremental predictions rather utilizes dynamic nature of spiking architecture. This demonstrates how well \textit{SpikeMS} generalize outside the temporal domain.






\subsection{Qualitative Results}

Fig. \ref{fig:DatasetEval} shows qualitative results of our approach on the two datasets. For each example  the input, moving object groundtruth, and network prediction are shown. Note that, we show the event projections for clarity purposes but the network input and outputs are the event cloud/spikes. We can observe that the network output predictions are similar to the ground truth for the moving objects in the presence of significant background variation and motion dynamics. 

\begin{figure}[t!]
\begin{center}
    \includegraphics[width=\columnwidth]{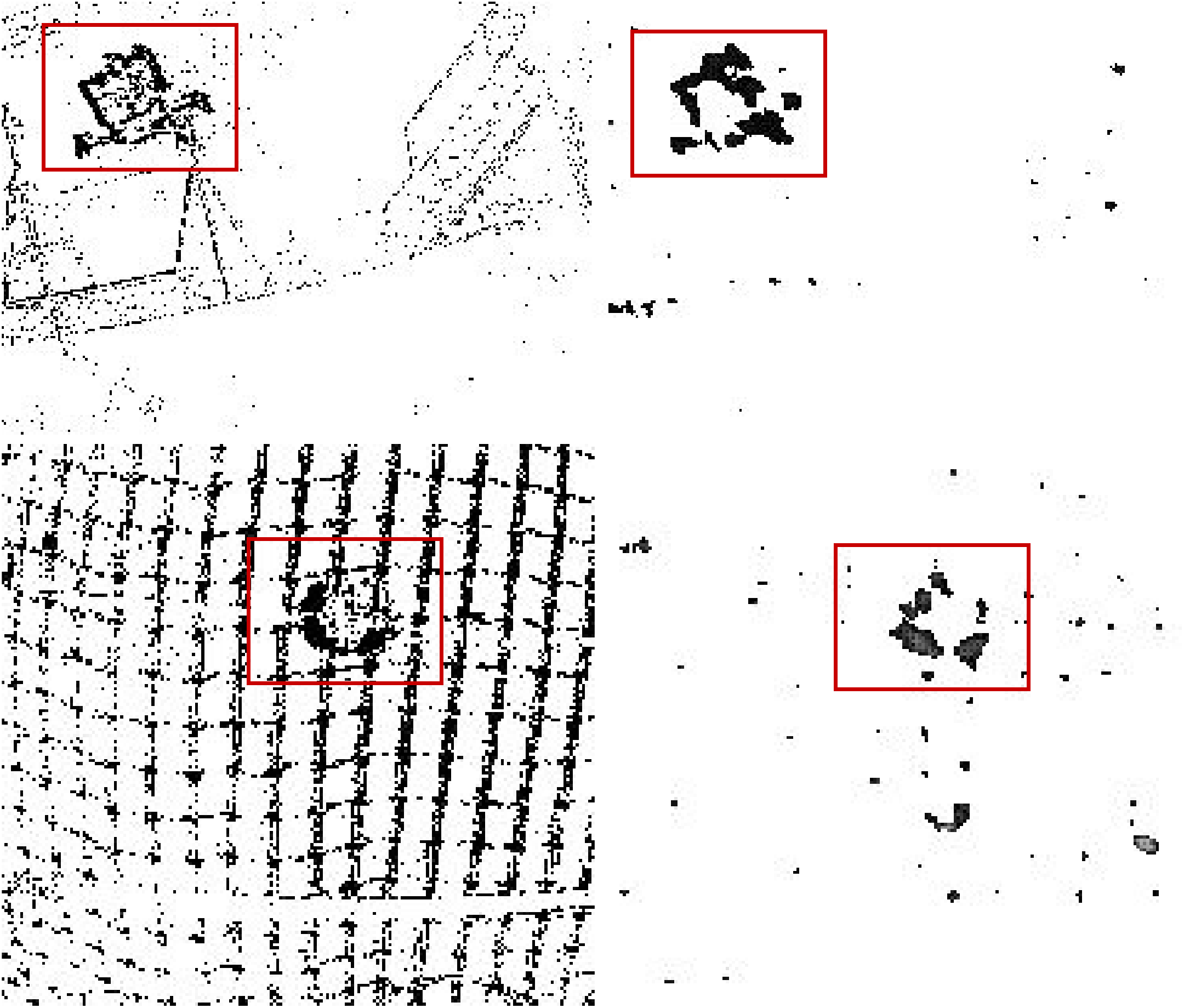}
\end{center}
\caption{Results showing motion segmentation generalization without fine-tuning or re-training on real world data.  \textit{SpikeMS} is able to segment the moving object from the scene even in the presence of substantial background noise.  The objects in the red bounding box are the true moving objects. Top row: A fast drone approaching a moving event camera. Bottom row: Moving object behind netted background.}

\label{fig:RealData}
\end{figure}

Fig. \ref{fig:RealData} shows the performance of \textit{SpikeMS} on real-world event streams, again with significant background variations and patterns. These results demonstrate the capability of \textit{SpikeMS} to generalize to different environments without any retraining or fine tuning of the network.

\subsection{Power Efficiency}

We further analyze the  benefits of \textit{SpikeMS}  compared to a fully ANN architecture with respect to power consumption. It is important to note that the main power consumption benefits occur when the SNN is implemented on a neuromorphic hardware such as the Intel$^{\text{\textregistered}}$ Loihi \cite{davies2018loihi}, where the network \textit{only} consumes power when there is a spike. Hence, the power consumption depends on the mean spike activity of the incoming data and the number of synaptic operations. In contrast,  ANNs perform dense matrix operations without exploiting the event sparsity. Thus, in anticipation of deploying \textit{SpikeMS} on the new generation of neurmorphic chips, we demonstrate the power savings by comparing the number of operations by a metric proposed in \cite{lee2020spike}.

 Table \ref{tab:power} provides the average number of synaptic operations in SNNs  along with a conservative estimate of the  energy benefit  when compared to an ANN-2D. We can observe that SNNs require a significantly lower number of synaptic operations and power as compared to ANNs. 

 
 \begin{table}
    \centering
    \caption {Comparison with state-of-the-art classical approaches for EED, MOD, EV-IMO datasets.}
    \label{tab:DetectionRateResults}
    \begin{tabular}{cccc}
    \hline
    \multirow{2}{*}{Method} & \multicolumn{3}{c}{Detection Rate (\%) $\uparrow$}\\
         & EED & MOD & EV-IMO\\
         \hline
         Mitrokhin \etal \cite{mitrokhin2018event}  & 88.93 & 70.12 & 48.79\\
          Stoffregen \etal \cite{stoffregen2019event} &  93.17 & - & - \\
          0-MMS\cite{parameshwara2021moms} & \textbf{94.2} & \textbf{82.35} & \textbf{81.06}\\
          \hline
           Ours & \textbf{91.5} & \textbf{68.82} & \textbf{65.14}\\
         \hline
    \end{tabular}
\end{table}

\begin{table}
    \centering
    \caption {Performance Metrics for EV-IMO and MOD datasets.}
    \label{tab:power}
    \resizebox{\columnwidth}{!}{%
    \begin{tabular}{ccccccc}
    \hline
        \multirow{2}{*}{Method} & \multicolumn{5}{c}{EV-IMO} &  \multicolumn{1}{c}{MOD}\\
             & \texttt{boxes}
             & \texttt{floor}
             & \texttt{wall} 
             & \texttt{table}
             & \texttt{fast}  
             & \texttt{}\\
        \hline
        Num. Operations ($\times 10^8$) & 0.42 & 0.34 & 0.52 & 0.38 & 0.53 & 0.83 \\
       Energy benefit ($\times$) & 116.19 & 143.53 & 93.84 & 128.42 & 92.07 & 58.80\\
    \hline
    \end{tabular}%
    }
\end{table}




\section{Conclusion}

We presented a first deep encoder-decoder Spiking Neural Network for a large-scale problem and demonstrated our architecture on the task of motion segmentation using data from a monocular event camera. Our novel spatio-temporal loss formulation takes full advantage of the  spatio-temporal nature of the event data. We demonstrated the unique ability of our network, \textit{SpikeMS}, for incremental prediction and showed its capability to generalize  across a range of temporal intervals without explicit augmentation. A comprehensive qualitative and quantitative evaluation was provided using  synthetic and real-world sequences from the EV-IMO, EED and MOD datasets. It was shown that \textit{SpikeMS} achieves   performance comparable  to an ANN method, but with 50$\times$ less power consumption. 

\section*{Acknowledgement}
The support of the National Science Foundation under grants BCS 1824198 and OISE 2020624 and the support of the Office of Naval Research under grant award N00014-17-1-2622 are gratefully acknowledged. We also would like to thank Samsung for providing us with the event-based vision sensor used in this study.

\bibliographystyle{./bibliography/IEEEtran}
\bibliography{./bibliography/IEEEabrv,./bibliography/IEEEexample}

\begin{thebibliography}{10}
\providecommand{\url}[1]{#1}
\csname url@samestyle\endcsname
\providecommand{\newblock}{\relax}
\providecommand{\bibinfo}[2]{#2}
\providecommand{\BIBentrySTDinterwordspacing}{\spaceskip=0pt\relax}
\providecommand{\BIBentryALTinterwordstretchfactor}{4}
\providecommand{\BIBentryALTinterwordspacing}{\spaceskip=\fontdimen2\font plus
\BIBentryALTinterwordstretchfactor\fontdimen3\font minus
  \fontdimen4\font\relax}
\providecommand{\BIBforeignlanguage}[2]{{%
\expandafter\ifx\csname l@#1\endcsname\relax
\typeout{** WARNING: IEEEtran.bst: No hyphenation pattern has been}%
\typeout{** loaded for the language `#1'. Using the pattern for}%
\typeout{** the default language instead.}%
\else
\language=\csname l@#1\endcsname
\fi
#2}}
\providecommand{\BIBdecl}{\relax}
\BIBdecl

\bibitem{diehl2015fast}
P.~U. Diehl, D.~Neil, J.~Binas, M.~Cook, S.-C. Liu, and M.~Pfeiffer,
  ``Fast-classifying, high-accuracy spiking deep networks through weight and
  threshold balancing,'' in \emph{2015 International Joint Conference on Neural
  Networks (IJCNN)}, 2015, pp. 1--8.

\bibitem{pfeiffer2018deep}
M.~Pfeiffer and T.~Pfeil, ``Deep learning with spiking neurons: opportunities
  and challenges,'' \emph{Frontiers in Neuroscience}, vol.~12, p. 774, 2018.

\bibitem{sanket2019evdodgenet}
N.~J. Sanket, C.~M. Parameshwara, C.~D. Singh, A.~V. Kuruttukulam,
  C.~Ferm\"uller, D.~Scaramuzza, and Y.~Aloimonos, ``{EVDodgeNet:} deep dynamic
  obstacle dodging with event cameras,'' 2019.

\bibitem{gehrig2020event}
M.~Gehrig, S.~B. Shrestha, D.~Mouritzen, and D.~Scaramuzza, ``Event-based
  angular velocity regression with spiking networks,'' in \emph{2020 IEEE
  International Conference on Robotics and Automation (ICRA)}, 2020, pp.
  4195--4202.

\bibitem{zenke2018superspike}
F.~Zenke and S.~Ganguli, ``Superspike: Supervised learning in multilayer
  spiking neural networks,'' \emph{Neural Computation}, vol.~30, no.~6, pp.
  1514--1541, 2018.

\bibitem{shrestha2018slayer}
S.~B. Shrestha and G.~Orchard, ``Slayer: Spike layer error reassignment in
  time,'' \emph{arXiv preprint arXiv:1810.08646}, 2018.

\bibitem{davies2018loihi}
M.~Davies, N.~Srinivasa, T.-H. Lin, G.~Chinya, Y.~Cao, S.~H. Choday, G.~Dimou,
  P.~Joshi, N.~Imam, S.~Jain \emph{et~al.}, ``Loihi: A neuromorphic manycore
  processor with on-chip learning,'' \emph{IEEE Micro}, vol.~38, no.~1, pp.
  82--99, 2018.

\bibitem{akopyan2015truenorth}
F.~Akopyan, J.~Sawada, A.~Cassidy, R.~Alvarez-Icaza, J.~Arthur, P.~Merolla,
  N.~Imam, Y.~Nakamura, P.~Datta, G.-J. Nam \emph{et~al.}, ``Truenorth: Design
  and tool flow of a 65 mw 1 million neuron programmable neurosynaptic chip,''
  \emph{IEEE Transactions on Computer-Aided Design of Integrated Circuits and
  Systems}, vol.~34, no.~10, pp. 1537--1557, 2015.

\bibitem{lichtsteiner2008128}
P.~Lichtsteiner, C.~Posch, and T.~Delbruck, ``A $128 \times 128$, 120 db 15
  $\mu$ s latency asynchronous temporal contrast vision sensor,'' \emph{IEEE
  Journal of Solid-State Circuits}, vol.~43, no.~2, pp. 566--576, 2008.

\bibitem{posch2010qvga}
C.~Posch, D.~Matolin, and R.~Wohlgenannt, ``A qvga 143 db dynamic range
  frame-free pwm image sensor with lossless pixel-level video compression and
  time-domain cds,'' \emph{IEEE Journal of Solid-State Circuits}, vol.~46,
  no.~1, pp. 259--275, 2010.

\bibitem{hodgkin1952quantitative}
A.~L. Hodgkin and A.~F. Huxley, ``A quantitative description of membrane
  current and its application to conduction and excitation in nerve,''
  \emph{The Journal of Physiology}, vol. 117, no.~4, pp. 500--544, 1952.

\bibitem{sejnowski1989hebb}
T.~J. Sejnowski and G.~Tesauro, ``The {Hebb} rule for synaptic plasticity:
  algorithms and implementations,'' in \emph{Neural Models of
  Plasticity}.\hskip 1em plus 0.5em minus 0.4em\relax Elsevier, 1989, pp.
  94--103.

\bibitem{nessler2009stdp}
B.~Nessler, M.~Pfeiffer, and W.~Maass, ``Stdp enables spiking neurons to detect
  hidden causes of their inputs,'' \emph{Advances in neural information
  processing systems}, vol.~22, pp. 1357--1365, 2009.

\bibitem{kheradpisheh2018stdp}
S.~R. Kheradpisheh, M.~Ganjtabesh, S.~J. Thorpe, and T.~Masquelier,
  ``Stdp-based spiking deep convolutional neural networks for object
  recognition,'' \emph{Neural Networks}, vol.~99, pp. 56--67, 2018.

\bibitem{evanusa2020deep}
M.~Evanusa, C.~Ferm\"uller, and Y.~Aloimonos, ``A deep 2-dimensional dynamical
  spiking neuronal network for temporal encoding trained with {STDP},''
  \emph{arXiv preprint arXiv:2009.00581}, 2020.

\bibitem{lee2018training}
C.~Lee, P.~Panda, G.~Srinivasan, and K.~Roy, ``Training deep spiking
  convolutional neural networks with stdp-based unsupervised pre-training
  followed by supervised fine-tuning,'' \emph{Frontiers in Neuroscience},
  vol.~12, p. 435, 2018.

\bibitem{lee2016training}
J.~H. Lee, T.~Delbruck, and M.~Pfeiffer, ``Training deep spiking neural
  networks using backpropagation,'' \emph{Frontiers in neuroscience}, vol.~10,
  p. 508, 2016.

\bibitem{neil2016learning}
D.~Neil, M.~Pfeiffer, and S.-C. Liu, ``Learning to be efficient: Algorithms for
  training low-latency, low-compute deep spiking neural networks,'' in
  \emph{Proceedings of the 31st Annual ACM Symposium on Applied Computing},
  2016, pp. 293--298.

\bibitem{orchard2015converting}
G.~Orchard, A.~Jayawant, G.~K. Cohen, and N.~Thakor, ``Converting static image
  datasets to spiking neuromorphic datasets using saccades,'' \emph{Frontiers
  in Neuroscience}, vol.~9, p. 437, 2015.

\bibitem{orchard2015hfirst}
G.~Orchard, C.~Meyer, R.~Etienne-Cummings, C.~Posch, N.~Thakor, and
  R.~Benosman, ``Hfirst: A temporal approach to object recognition,''
  \emph{{IEEE Transactions on Pattern Analysis and Machine Intelligence}},
  vol.~37, no.~10, pp. 2028--2040, 2015.

\bibitem{amir2017low}
A.~Amir, B.~Taba, D.~Berg, T.~Melano, J.~McKinstry, C.~Di~Nolfo, T.~Nayak,
  A.~Andreopoulos, G.~Garreau, M.~Mendoza \emph{et~al.}, ``A low power, fully
  event-based gesture recognition system,'' in \emph{Proceedings of the IEEE
  Conference on Computer Vision and Pattern Recognition}, 2017, pp. 7243--7252.

\bibitem{haessig2018}
G.~Haessig, A.~Cassidy, R.~Alvarez, and G.~Benosman, R.and~Orchard, ``Spiking
  optical flow for event-based sensors using {IBM’s TrueNorth} neurosynaptic
  system,'' pp. 860--870, 2018.

\bibitem{paredes2019unsupervised}
F.~Paredes-Vall{\'e}s, K.~Y. Scheper, and G.~C. de~Croon, ``Unsupervised
  learning of a hierarchical spiking neural network for optical flow
  estimation: From events to global motion perception,'' \emph{IEEE
  Transactions on Pattern Analysis and Machine Intelligence}, vol.~42, no.~8,
  pp. 2051--2064, 2019.

\bibitem{haessig2019spiking}
G.~Haessig, X.~Berthelon, S.-H. Ieng, and R.~Benosman, ``A spiking neural
  network model of depth from defocus for event-based neuromorphic vision,''
  \emph{Scientific reports}, vol.~9, no.~1, pp. 1--11, 2019.

\bibitem{renner2019event}
A.~Renner, M.~Evanusa, and Y.~Sandamirskaya, ``Event-based attention and
  tracking on neuromorphic hardware,'' in \emph{2019 IEEE/CVF Conference on
  Computer Vision and Pattern Recognition Workshops (CVPRW)}, 2019, pp.
  1709--1716.

\bibitem{lee2020spike}
C.~Lee, A.~K. Kosta, A.~Z. Zhu, K.~Chaney, K.~Daniilidis, and K.~Roy,
  ``Spike-flownet: event-based optical flow estimation with energy-efficient
  hybrid neural networks,'' in \emph{European Conference on Computer
  Vision}.\hskip 1em plus 0.5em minus 0.4em\relax Springer, 2020, pp. 366--382.

\bibitem{barranco2018real}
F.~Barranco, C.~Ferm\"uller, and E.~Ros, ``Real-time clustering and
  multi-target tracking using event-based sensors,'' in \emph{2018 IEEE/RSJ
  International Conference on Intelligent Robots and Systems (IROS)}, 2018, pp.
  5764--5769.

\bibitem{mitrokhin2018event}
A.~Mitrokhin, C.~Ferm{\"u}ller, C.~Parameshwara, and Y.~Aloimonos,
  ``Event-based moving object detection and tracking,'' in \emph{2018 IEEE/RSJ
  International Conference on Intelligent Robots and Systems (IROS)}.\hskip 1em
  plus 0.5em minus 0.4em\relax IEEE, 2018, pp. 1--9.

\bibitem{stoffregen2019event}
T.~Stoffregen, G.~Gallego, T.~Drummond, L.~Kleeman, and D.~Scaramuzza,
  ``Event-based motion segmentation by motion compensation,'' in
  \emph{Proceedings of the IEEE/CVF International Conference on Computer
  Vision}, 2019, pp. 7244--7253.

\bibitem{parameshwara2021moms}
C.~M. Parameshwara, N.~J. Sanket, C.~Deep~Singh, C.~Ferm\"uller, and
  Y.~Aloimonos, ``0-mms: Zero-shot multi-motion segmentation with a monocular
  event camera,'' in \emph{IEEE International Conference on Robotics and
  Automation (ICRA)}, 2021.

\bibitem{mitrokhin2019ev}
A.~Mitrokhin, C.~Ye, C.~Ferm{\"u}ller, Y.~Aloimonos, and T.~Delbruck,
  ``{EV-IMO:} motion segmentation dataset and learning pipeline for event
  cameras,'' in \emph{IEEE/RSJ International Conference on Intelligent Robots
  and Systems (IROS)}, 2019.

\bibitem{mitrokhin2020learning}
A.~Mitrokhin, Z.~Hua, C.~Ferm\"uller, and Y.~Aloimonos, ``Learning visual
  motion segmentation using event surfaces,'' in \emph{Proceedings of the
  IEEE/CVF Conference on Computer Vision and Pattern Recognition}, 2020, pp.
  14\,414--14\,423.

\bibitem{gerstner1995time}
W.~Gerstner, ``Time structure of the activity in neural network models,''
  \emph{Physical review E}, vol.~51, no.~1, p. 738, 1995.

\bibitem{Gerstner:2008}
------, ``{S}pike-response model,'' \emph{Scholarpedia}, vol.~3, no.~12, p.
  1343, 2008, revision \#91800.

\bibitem{vanRossum}
M.~van Rossum, ``A novel spike distance,'' \emph{Neural Computation}, vol.~13,
  pp. 751--763, 04 2001.

\bibitem{Kreuz:2011}
T.~Kreuz, ``{M}easures of spike train synchrony,'' \emph{Scholarpedia}, vol.~6,
  no.~10, p. 11934, 2011, revision \#190333.

\bibitem{qi2017pointnet}
C.~R. Qi, L.~Yi, H.~Su, and L.~J. Guibas, ``Pointnet++: Deep hierarchical
  feature learning on point sets in a metric space,'' 2017.

\bibitem{kalman1960new}
R.~Kalman, ``A new approach to linear filtering and prediction problems,''
  \emph{Journal of Basic Engineering}, vol.~82, no.~1, pp. 35--45, 1960.

\end{thebibliography}

\end{document}